\documentclass{article}
\usepackage[preprint]{neurips_2022}
\usepackage[latin1]{inputenc}
\usepackage{graphicx}
\usepackage{amsmath}
\usepackage{amsfonts}
\usepackage{amssymb}
\usepackage{makeidx}
\usepackage{setspace}
\usepackage{geometry}
\usepackage{color}
\usepackage{hyperref}
\usepackage{natbib}
\usepackage{epigraph} 
\usepackage{caption} 
\usepackage[table]{xcolor}
\title{How much human-like visual experience do current self-supervised learning algorithms need in order to achieve human-level object recognition?}
\author{%
  Emin Orhan \\
  New York University\\
  \texttt{aeminorhan@gmail.com}
  }
\date{}
\begin{document}

\maketitle

\begin{abstract}
This paper addresses a fundamental question: how good are our current self-supervised visual representation learning algorithms relative to humans? More concretely, how much ``human-like'' natural visual experience would these algorithms need in order to reach human-level performance in a complex, realistic visual object recognition task such as ImageNet? Using a scaling experiment, here we estimate that the answer is several orders of magnitude longer than a human lifetime: typically on the order of a million to a billion years of natural visual experience (depending on the algorithm used). We obtain even larger estimates for achieving human-level performance in ImageNet-derived robustness benchmarks. The exact values of these estimates are sensitive to some underlying assumptions, however even in the most optimistic scenarios they remain orders of magnitude larger than a human lifetime. We discuss the main caveats surrounding our estimates and the implications of these surprising results.
\end{abstract}

\section{Introduction}

\epigraph{\textit{When you cannot express it in numbers, your knowledge is of a meagre and unsatisfactory kind.}}{--- William Thomson}

Deep learning models are often trained with large amounts of data of a kind that is quite unnatural from a human perspective: e.g.~millions of static images downloaded from the internet for training image recognition models or the entire text data on the internet for training language models. The resulting systems have impressive capabilities certainly \citep{brown2020,ramesh2021,ramesh2022}, but this mismatch between the kind and amount of data they are trained with and the kind and amount of data humans need to develop similar capabilities often invites the criticism that these models are possibly not as sample efficient as humans and hence not as sample efficient as they could potentially be if only we hardcoded more human-like inductive biases in them \citep{lake2017,marcus2020}. 

However, the only reliable way to ascertain this claim would be to simulate as faithfully as possible the full sensorium of a developing human being and see how far we can get in terms of capabilities with our current deep learning models and algorithms applied to it. This is presumably currently not feasible for technical reasons (possibly also for more fundamental conceptual reasons). But we can still run more limited versions of this kind of experiment even today and hope to obtain meaningful results. For example, if we were to fully record in video the visual experiences of a developing human being during the course of several years of their life and apply our state-of-the-art machine learning models and algorithms to these video recordings, what kind of a system would we be able to get in terms of perceptual capabilities? Would the system approach human-level competence in practically important perceptual tasks such visual object recognition or would it fall far short of it? The answer to this question would have important implications: the first possibility would indicate that our current algorithms are at least as sample efficient as humans (at least when it comes to particular capabilities such as visual object recognition) even when ignoring the important additional inputs humans receive due to embodiment. The second possibility would lead us to consider integrating additional inputs and/or inductive biases to our current models/algorithms to make them more sample efficient. 

In this work, we carry out an experiment of exactly this sort based on a dataset of $\sim$1301 hours ($\sim$54 days) of ``natural'' video. 54 days is, of course, still quite short compared to the years of natural visual experience that a typical, sighted human being would receive over the course of their development, so we develop a scaling argument to estimate the amount of ``natural'' visual data of this sort it would take current state-of-the-art self-supervised representation learning algorithms to reach human-level performance in a complex, realistic object recognition benchmark, namely ImageNet. 

We borrow our estimate of the human-level performance on ImageNet from \citet{russakovsky2015}, where a psychophysical experiment with human subjects was performed in order to estimate their top-5 accuracy on the ImageNet test set. Replicating the human experimental design in \citet{russakovsky2015} \textit{in silico} (i.e.~finetuning the self-supervised model trained with natural video only with $\sim$1\% of the ImageNet training data), we estimate that it would take current state-of-the-art self-supervised learning algorithms on the order of a million to a billion years of natural video data to reach human-level performance on ImageNet, depending on the particular algorithm used. This estimate is orders of magnitude larger than the typical human lifetime.

When we take into account not only the accuracy but also the robustness of human object recognition, as demonstrated and rigorously quantified in ImageNet-derived out-of-distribution generalization benchmarks \citep{geirhos2021}, we find even larger estimates for the amount of natural visual experience it would take current self-supervised learning algorithms to achieve human-level object recognition. We discuss the implications of these findings, as well as some of the main caveats attached to our estimates.\footnote{Code, stimuli, and pretrained models are available at: \url{https://github.com/eminorhan/human-ssl}.}
\vspace{-1em}
\section{Methods}
\vspace{-1em}
\subsection{Datasets}
\label{sec_datasets}
We combined five video datasets for self-supervised representation learning. Our main criterion for inclusion was that the videos in the dataset be representative of the visual experiences of a typical human being (young or adult) in terms of content and style as much as possible. Two important properties of the natural human visual experience we focused on, in particular, were its long, continuous character and its egocentric nature. Therefore, standard video datasets in computer vision, such as Kinetics \citep{smaira2020}, which consist of a large number of short video clips of a few seconds long in duration, shot from a third-person perspective, are not suitable for our purposes.\footnote{The new Ego4d dataset \citep{grauman2021} was released during the preparation of this manuscript.} We instead selected the following five datasets. These datasets all consist of relatively long, continuous videos (usually at least on the order of tens of minutes long each) and all but one (AVA) are datasets of egocentric headcam recordings (the justification for including AVA, which is a dataset of movies, is that watching movies is arguably a common pastime in the modern world, hence movies can be plausibly claimed to be representative visual stimuli for humans):

\textbf{SAYCam}: SAYCam is a large, longitudinal, audiovisual dataset of headcam recordings collected from the perspective of three young children between the ages of 6 to 32 months \citep{sullivan2020}. The combined length of the videos from all three children is about 500 hours. The recordings were made approximately once a week for $\sim$1-2 hours (usually continuously) over the course of a $\sim$2.5 year period during the early development of the children. This dataset is privately hosted on the Databrary repository for behavioral science and is not publicly accessible, but researchers can apply for access.\footnote{For details, please see: \url{https://databrary.org/about/agreement.html}.}

\textbf{AVA (v2.2)}: AVA is a dataset of films. The dataset comes with rich annotations for parts of these films \citep{gu2018}. However, we use the entire films for self-supervised representation learning in this work, not just the annotated parts. Both \texttt{trainval} and \texttt{test} videos (299 and 131 videos respectively) are used for self-supervised training. The total length of the videos from this dataset is about 640 hours. The dataset is publicly accessible from: \url{https://research.google.com/ava/download.html}.

\textbf{KrishnaCam}: KrishnaCam is a dataset of headcam videos collected by a graduate student \citep{singh2016}. It contains long, continuous videos of daily episodes in the life of the graduate student. The total length of the videos in the dataset is about 70 hours. The dataset is publicly accessible from: \url{https://krsingh.cs.ucdavis.edu/krishna_files/papers/krishnacam/krishnacam.html}

\textbf{Epic-Kitchens}: This dataset contains relatively long headcam videos from multiple participants performing daily culinary chores in their kitchens \citep{damen2018}. The total length of the videos in this dataset is about 80 hours. The dataset is publicly accessible from: \url{https://epic-kitchens.github.io/2021}.

\textbf{UT Ego}: UT Ego contains four continuous headcam recorded videos, each 3-5 hours long \citep{lee2012}. The dataset is publicly accessible from: \url{http://vision.cs.utexas.edu/projects/egocentric_data/UT_Egocentric_Dataset.html}.

As mentioned before, the total length of the combined dataset is $\sim$1301 hours ($\sim$54 days). The videos are temporally subsampled at a rate of 5 frames per second, which results in a combined dataset with $\sim$23M frames in total. The frames are resized and cropped in the standard ImageNet format: first resized so that the minor edge is 256 pixels long and then a central crop of size 224$\times$224 taken from the resized frame. 

\subsection{Self-supervised learning algorithms}
We used two different self-supervised visual representation learning algorithms in our experiments: the \textit{temporal classification} algorithm introduced in \citet{orhan2020} for longitudinal video data and the \textit{DINO} algorithm introduced in \citet{caron2021} for image-based self-supervised representation learning. We used these two qualitatively different representation learning algorithms in our experiments to make sure our overall conclusions are not very sensitive to the choice of the self-supervised learning algorithm. We now briefly describe these algorithms (we refer the reader to the original papers for further details about the respective algorithms). 

\textbf{Temporal classification:} The high level idea in temporal classification is to extract useful high level visual representations from a scene by learning temporal invariances \citep{wiskott2002}. The particular way in which this is implemented in temporal classification is to divide the entire combined dataset (end-to-end concatenated) into relatively short temporal episodes and set up the self-supervised learning objective as a standard classification task where the goal is to predict the temporal episode label for any given frame (``which episode does this frame come from?''). 

For our implementation of the temporal classification algorithm, we use the public code base provided by the authors.\footnote{Available at: \url{https://github.com/eminorhan/baby-vision}} Most of the hyperparameter choices in our algorithm are borrowed from \citet{orhan2020}. In particular, we use an episode length of 288 seconds, slightly shorter than 5 minutes, which divides the entire $\sim$1.3k hour dataset into a total of 16279 classes or episodes. The main difference from \citet{orhan2020} is that we use a different data augmentation pipeline here with stronger augmentations, similar to those used in SimCLR \citep{chen2020}. A full description of the data augmentation pipeline, as well as other training details can be found in the Appendix.


\textbf{DINO:} DINO is a self-distillation based representation learning algorithm \citep{caron2021}. At a high level, it is similar to other distillation-based self-supervised learning algorithms such as Noisy Student \citep{xie2020} or BYOL \citep{grill2020}. In DINO, a student network and a teacher network receive differently augmented versions of the same image and the learning objective is to match the final image representation in the student network to that in the teacher network. During training, the parameters of the teacher network are updated as a moving average of the student network's parameters. Unlike in other self-distillation based approaches, DINO uses the exact same architecture for the student and the teacher networks (but with different parameters) and it also uses a cross-entropy objective to measure the representational dissimilarity between the student and the teacher networks. 

Intuitively, the algorithm works because the two networks receive different copies of the same image, hence the teacher network is always able to impart new information to the student network and, as a result, improve it. Because the teacher itself is simply a moving average of the student, improvements in the student network automatically transfer to the teacher network in later iterations. 

DINO is a highly competitive algorithm for self-supervised representation learning from images, achieving close to state-of-the-art results on ImageNet. Here, we apply DINO to video data by treating the video frames as independent images. Thus, unlike the temporal classification algorithm, DINO does not explicitly exploit the temporal structure of the data.

For our implementation of DINO, we use the code base made public by the authors.\footnote{Available at: \url{https://github.com/facebookresearch/dino}} We generally adopt the default hyperparameter settings provided by the authors with minor modifications. A full description of the hyperparameter values and other training details can be found in the Appendix.    

For both algorithms, our computational resources did not allow us to carry out large, systematic hyperparameter searches. We instead made our hyperparameter choices based on a small amount of manual search around values/settings motivated by previous works. It is possible that more extensive hyperparameter searches, especially as regards the data augmentation policies, could improve some of the results reported below.  
\vspace{-1em}
\subsection{Models}
For the models trained with the temporal classification algorithm, we used a \texttt{resnext101\_32x8d} backbone (available from \texttt{torchvision.models}). This model has a total size of $\sim$88M parameters (with the ImageNet head) and an embedding layer with 2048 dimensions. When fully trained on ImageNet (with labels), it achieves a respectable top-1 accuracy of 79.3\% and a top-5 accuracy of 94.5\% on the validation set. 

For the models trained with the DINO algorithm, we used a smaller \texttt{resnext50\_32x4d} backbone. We had to use a smaller backbone in this case, since this algorithm has larger memory and compute requirements because of the existence of two separate networks, the student and the teacher. The \texttt{resnext50\_32x4d} model has roughly 25M parameters (with the ImageNet head) and an embedding layer of size 2048. When fully trained on ImageNet (with labels), it achieves a top-1 accuracy of 77.6\% and a top-5 accuracy of 93.7\% on the validation set. 

With both algorithms, the models were trained up to convergence on the natural video data. On the entire dataset, consisting of $\sim$23M video frames as described above, training took roughly two weeks to complete in each case, using data parallelism on 4 NVIDIA V100 GPUs (we used the largest batch size we could fit into the GPUs in both cases).

\section{Results}
\subsection{How much human-like visual experience do current self-supervised learning algorithms need in order to reach human-level object recognition \textit{accuracy} on ImageNet?}
We take our estimate of the human-level performance on ImageNet from \citet{russakovsky2015}. They report the top-5 classification error of two trained human annotators as 5.1\% and 12.0\%, using a few-shot learning type experimental design where the annotators had access to $\sim$1\% of the training data with labels (13 images from each of the 1000 classes). Since both of these estimates are based on stochastic samples from the test set, instead of being unnecessarily precise here, we will use 10\% top-5 error (or 90\% top-5 accuracy) as a plausible lower bound for human-level performance on ImageNet and report our estimates with respect to this performance level below. Using the more stringent criterion of 5\% top-5 error instead typically increases our estimates by at least an order of magnitude.

We replicate the few-shot learning experimental setup in \citet{russakovsky2015} for our self-supervised models by finetuning them with 13000 randomly sampled labeled images from the ImageNet training set. Importantly, in order to extrapolate the behavior of the self-supervised models beyond the $\sim$1.3k hours of natural video data we have, we also repeat this experiment for self-supervised models trained with 10\%, 1\%, and 0.1\% of the self-supervised training data (sampled as a temporally contiguous block in each case), thus covering four orders of magnitude in the amount of natural video data used for self-supervised learning. Since this subset selection is a stochastic process, we repeat each experiment 3 times.

\begin{figure}
  \centering
    \includegraphics[width=1.0\textwidth, trim=0mm 0mm 0mm 0mm, clip]{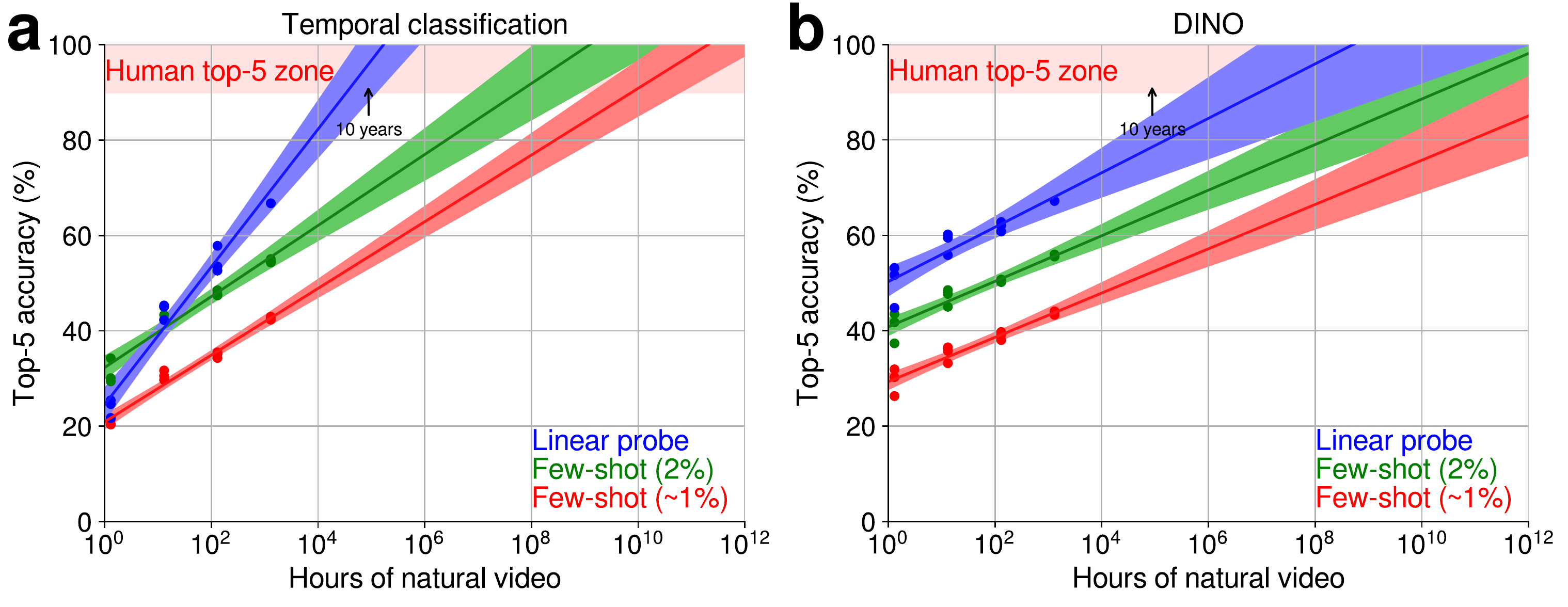}
  \caption{Top-5 validation accuracy on ImageNet as a function of the amount of natural video data used for self-supervised learning with the temporal classification (a) and DINO (b) algorithms.~Results with three different evaluation protocols are shown: linear probing (blue), few-shot learning with $\sim$1\% of the labeled training data (red) as in the human experiments in \citet{russakovsky2015}, and few-shot learning with 2\% of the labeled training data (green) to give an indication of the sensitivity of our few-shot estimates. Based on \citet{russakovsky2015}, we consider accuracies above 90\% as human-level performance.~Each dot represents a different run (the exact values of these data points are documented in the Appendix).~The straight lines are linear fits, the shaded regions represent 95\% confidence intervals around the linear fit. As a reference, we also indicate the 10 year mark, which is a meaningful time scale for human visual development, with a vertical black arrow in each plot.}
  \label{scaling_fig}
\end{figure}

Figure~\ref{scaling_fig} presents our main result.~We linearly extrapolate the top-5 accuracy on the ImageNet validation set as a function of the logarithm of the hours of natural video used during self-supervised learning.\footnote{For completeness, we also show the scaling of the top-1 accuracies in each condition in the Appendix, however these will not be relevant for the main objective of this paper.}~In the few-shot learning setup used for the human experiments in \citet{russakovsky2015} (with $\sim$1\% of the labeled training data), we get very large estimates for the amount of human-like natural video necessary to reach 90\% top-5 accuracy (red lines in Figure~\ref{scaling_fig}): on the order of a million years for temporal classification and on the order of a billion years for DINO. These estimates are documented in Table~\ref{scaling_tab}. 

\begin{table}
\centering
\begin{tabular}{ccccc}
 \texttt{eval.~method} & \texttt{Temporal classification} & \texttt{DINO} \\
 \hline
 \texttt{few-shot ($\sim$1\%)}  &  \cellcolor{red!25}0.9M \footnotesize(0.2M, 6.7M) & \cellcolor{red!25}1.4G \footnotesize(30.0M, 0.3T) \\
 \texttt{few-shot (2\%)}  &  \cellcolor{green!25} 6.4k \footnotesize(0.9k, 90.9k) & \cellcolor{green!25} 2.3M \footnotesize(0.1M, 0.2G) \\
\texttt{linear probe}  & 3.9  \footnotesize(1.5, 13.6) & 1.0k \footnotesize(46.5, 0.3M) \\
\hline
\end{tabular}
\caption{Amount of time (in \textbf{\textit{\underline{years}}}) it would take the self-supervised models to reach 90\% top-5 validation accuracy on ImageNet. Note that these estimates do not take sleep into account (i.e.~it is assumed that the models receive visual experience 24 hours/day). To factor in an 8 hours/day sleep period during which no visual stimulation is experienced, simply multiply these estimates by 1.5. The most relevant estimates (using the few-shot learning evaluation with $\sim$1\% or 2\% of the labeled data) are highlighted with a red and a green background, respectively. Numbers in parenthesis indicate the 95\% confidence intervals (the confidence intervals are not symmetric, since the linear fits are carried out in the log domain). Metric suffixes: k=$10^3$, M=$10^6$, G=$10^9$, T=$10^{12}$. Note that the age of Earth is roughly 4.5G years and our species, \textit{homo sapiens}, is roughly 0.3M years old.}
\label{scaling_tab}
\vspace{-2.5em}
\end{table}

The scaling difference between the two algorithms is mainly driven by the inferior performance of the temporal classification algorithm with the smaller data sizes. Both algorithms perform similarly at the largest data size (i.e.~with 1301 hours of natural video): temporal classification achieves a mean top-5 accuracy of 42.6\% and DINO achieves a mean top-5 accuracy of 43.7\% in this condition. However, DINO performs markedly better for smaller data sizes, e.g.~close to $\sim$9\% better with only 1.3 hours of natural video. This leads to shallower linear fits for DINO, hence larger estimates for the amount of natural video data needed to achieve a given level of accuracy on ImageNet.

A potential discrepancy between the human experiments in \citet{russakovsky2015} and our \textit{in silico} experiments is that the labeled examples our models see during few-shot learning are the \textit{only} labeled examples they experience from the ImageNet categories. However, this is very likely an unrealistic assumption for the human subjects: they had likely seen other labeled examples from at least some of the ImageNet categories (especially from the more common categories) prior to the experiment. This puts the models at a disadvantage relative to humans. 

To partially account for this potential discrepancy and to get a sense of the sensitivity of our estimates to the details of the evaluation protocol, we thus repeated our few-shot learning experiments with roughly twice as many labeled examples, i.e. using 2\% of the ImageNet training data this time. The results are show in green in Figure~\ref{scaling_fig}: doubling the amount of labeled examples shown during few-shot learning leads to 2-3 orders of magnitude smaller estimates (Table~\ref{scaling_tab}). This suggests that our estimates might be quite sensitive to the details of the evaluation protocol, however note that the estimates are still several orders of magnitude larger than developmentally realistic time scales for humans. We discuss the implications of this result in more detail in the next section.

Although few-shot learning is often used as an evaluation method for self-supervised learning algorithms, an even more common evaluation method is linear probing, where a single linear layer is trained (using the entire training data of the evaluation task) on top of the frozen embedding layer features that are learned exclusively through self-supervised learning. Linear probing evaluation is clearly not relevant for human comparisons on ImageNet, since the human experiments in \citet{russakovsky2015} were instead performed through few-shot learning, as discussed above (it is not even clear what a linear probe would mean in the human context). However, given the popularity of linear probing evaluations in the self-supervised learning literature, we were curious to see how the performance of our self-supervised model would scale under this evaluation measure. 

The results are shown in blue in Figure~\ref{scaling_fig} and estimates for the amount of natural visual experience necessary to reach 90\% top-5 accuracy under this evaluation protocol are documented in Table~\ref{scaling_tab}. Overall, we obtain much better performance and significantly smaller estimates for linear probing than for the few-shot evaluation protocols. With the DINO algorithm, the data scaling for linear probing is qualitatively similar to the data scaling for few-shot evaluations, but shifted upward (Figure~\ref{scaling_fig}b), whereas with the temporal classification algorithm, we observe a qualitatively different (and more favorable) data scaling behavior for linear probing vs.~few-shot evaluations (Figure~\ref{scaling_fig}a).

These results are consistent with prior literature on self-supervised visual representation learning and suggest that few-shot learning may be a much more stringent evaluation protocol than linear probing, especially when the data used for self-supervised learning differs from the evaluation data, as it does here. They also suggest that in a certain sense, ImageNet may be more ``linearly solvable'' than generally appreciated: a linear map constrained by the entire ImageNet training data may go a long way toward a decent level of performance on ImageNet, even on top of relatively weak features.

\vspace{0em}

\subsection{How much human-like visual experience do current self-supervised learning algorithms need in order to reach human-level \textit{robustness} in object recognition?}
A remarkable property of human object recognition is that it is not only highly accurate, but also highly robust to distortions, image manipulations, and other distributional shifts. State-of-the-art self-supervised learning algorithms trained on ImageNet can nowadays match or even surpass human-level accuracy on the clean ImageNet validation set. However, even models trained with large amounts of extra data often struggle to reach human-level accuracy on distorted versions of ImageNet designed to measure the out-of-distribution (OOD) generalization capabilities of these models \citep{geirhos2021}.  

\citet{geirhos2021} recently conducted a rigorous, head-to-head comparison between 90 human observers and a large variety of deep learning models across 17 different OOD generalization datasets derived from ImageNet, which we replicate here with our own self-supervised models trained on natural video data. These OOD generalization datasets include stylized, sketched, or silhouetted versions of the ImageNet validation images, as well as parametrically varied versions such as low-pass filtered or additively noised images. The full list can be found in the Appendix and further details regarding the stimuli and the human experiments can be found in \citet{geirhos2021}.    

\begin{figure}
  \centering
    \includegraphics[width=1.0\textwidth, trim=0mm 0mm 0mm 0mm, clip]{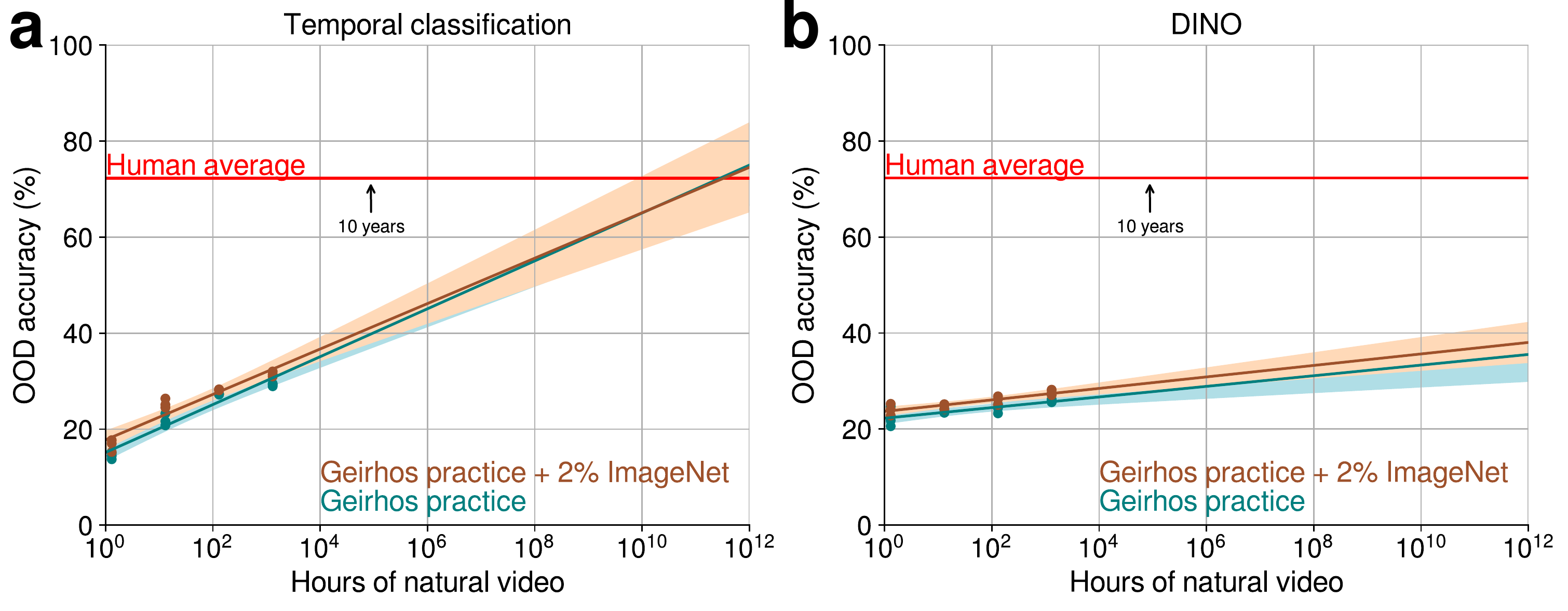}
  \caption{Average out-of-distribution (OOD) accuracy across 17 ImageNet-derived OOD generalization datasets as a function of the amount of natural video data used for self-supervised learning with the temporal classification (a) and DINO (b) algorithms. Models were either finetuned with the 321 practice images from \citet{geirhos2021} only (teal) or finetuned with both these practice images and 2\% of labeled data from the ImageNet training set in addition (brown). Each dot represents a different run (the exact values of these data points are provided in the Appendix).~The straight lines are linear fits, the shaded regions represent 95\% confidence intervals around the linear fit. As a reference, we also indicate the 10 year mark, which is a relevant time scale for human visual development, with a vertical black arrow in each plot. For comparison, we note that a supervised ResNet50 model trained on ImageNet achieves a mean OOD accuracy of 55.9\% on this benchmark.}
  \label{robustness_fig}
  \vspace{-1em}
\end{figure}

Because of the difficulty of performing experiments with humans involving 1000 different choices (as in ImageNet classification), \citet{geirhos2021} instead work with a reduced set of 16 basic-level categories in their experiments (these categories are listed in the Appendix). Before the experimental session, human subjects were given 320 practice trials (with supervised feedback) to familiarize them with the stimuli and the task, using a held-out dataset of images. To make a fair comparison between humans and our self-supervised models and also to teach our models the categorization task, we gave the same practice trials (with supervised feedback) to our self-supervised models trained with the natural video data (we do this by finetuning the entire model with the practice images and their labels). Similar to the previous scaling experiment, we then perform the same experiments with self-supervised models trained with 10\%, 1\%, and 0.1\% of the combined video data, respectively, then evaluate the OOD accuracy of the models on the \citet{geirhos2021} benchmarks, and then linearly extrapolate the performance. The results of this scaling experiment are shown in Figure~\ref{robustness_fig} and in Table~\ref{geirhos_tab}.

\begin{table}
\centering
\begin{tabular}{ccccc}
 \texttt{finetuned on} & \texttt{Temporal classification} & \texttt{DINO} \\
 \hline
\texttt{Geirhos practice only} & 32.9M \footnotesize(1.4M, 2.5G) &  $>$1T \footnotesize($>$1T, $>$1T) \\
\texttt{Geirhos practice + 2\% ImageNet} & 37.8M \footnotesize(1.0M, 6.9G) & $>$1T \footnotesize($>$1T, $>$1T) \\
\hline
\end{tabular}
\caption{Amount of time (in \textbf{\textit{\underline{years}}}) it would take the self-supervised models to reach human-level OOD accuracy on the \citet{geirhos2021} generalization benchmarks. These estimates again do not factor in sleep. Numbers in parenthesis indicate the 95\% confidence intervals (the confidence intervals are not symmetric, since the linear fits are carried out in the log domain). For DINO, the estimates were too large to be meaningful, as can be visually ascertained from Figure~\ref{robustness_fig}b.}
\label{geirhos_tab}
\vspace{-1em}
\end{table}

We find even larger estimates for the amount of natural video data it would take to reach human-level performance in these OOD generalization benchmarks than in the previous scaling experiment (Table~\ref{geirhos_tab}). The estimates for the temporal classification algorithm are smaller than those for DINO, but this is again largely driven by the relatively poor performance of the temporal classification algorithm for the smallest data size (1.3 hours of natural video). The estimates for DINO are too large to even be meaningful.

These estimates are again subject to the prior labeled example exposure problem discussed above in connection with our first scaling experiment. To address this issue and to find out the sensitivity of our estimates to the details of the finetuning procedure, we next performed the same experiments with self-supervised models finetuned with both the practice images from \citet{geirhos2021} as well as 2\% of the labeled examples from the ImageNet training set (in practice we do this by first finetuning on 2\% of ImageNet training data, removing the classifier head, and then finetuning on the \citet{geirhos2021} practice images with a new classifier head attached). Although this extra finetuning improves the OOD accuracy for both algorithms, the improvements are too small to make a material difference in the scaling behavior (compare the teal and brown linear fits in Figure~\ref{robustness_fig}). This suggests that unlike the scaling of clean validation accuracy, the scaling of OOD accuracy is not very sensitive to the finetuning details. 
\vspace{-1em}
\section{Caveats}
This work tackles an ambitious and difficult problem. Like most attempts at addressing a hard problem, it comes with a long list of caveats. Here we discuss some of the main caveats regarding our estimates of the amount of natural visual data current self-supervised learning algorithms would need in order to achieve human-level object recognition, with the hope that future work will address these caveats to come up with more accurate estimates. 

\textbf{1. Is the distribution of self-supervised training data realistic enough?} Although we tried to select our video datasets to be representative of the typical human visual experiences as much as possible, they obviously are not an unbiased sample from the visual experiences of any single individual. There are several good reasons to think that a more realistic, unbiased sample of the same size from a single individual would actually contain less variation, hence it would likely cause our estimates to be even larger. First, most of the individual video datasets in our sample consists of recordings from multiple individuals and, almost by definition, data from multiple individuals are expected to contain more variation than data of the same size from a single individual. Second, our sample contains data from multiple datasets recorded with different devices, which introduces additional variation. Third, a relatively large proportion of our sample (roughly half) consists of a large number films highly diverse in style and content, which likely contain a lot more variation than the typical daily visual experiences of a single individual. Fourth, we also used data augmentation during self-supervised learning, which introduces further variation to our sample. As much as possible, future work should try to use larger, more realistic, and more unbiased samples recorded from the perspective of a single individual and should prefer self-supervised learning algorithms that require less data augmentation: e.g.~likelihood-based generative self-supervised learning objectives tend to require less data augmentation \citep{chen2020b}.

\textbf{2. Is the linear extrapolation of performance appropriate?} We used a linear model to extrapolate the performance of our self-supervised model beyond the $\sim$1.3k hours of natural video data we had available. This provides a qualitatively good fit to the small number of data points we have (Figures~\ref{scaling_fig} and \ref{robustness_fig}), but will almost certainly start to break down especially as the performance approaches human-level performance on the ImageNet validation set and the related OOD generalization benchmarks. We expect that there will be diminishing returns from more data as this starts to happen, which would again lead to an increase in our estimates. On the other hand, a linear fit may underestimate the rate of improvement for the early parts of the curve, so it is difficult to guess what the net effect of a deviation from the linear trend might turn out to be. Ideally, future work should add more data points to Figures~\ref{scaling_fig} and \ref{robustness_fig} spanning several additional orders of magnitude so functions more flexible than a simple linear fit can be considered for the data size vs.~accuracy curves. 

\textbf{3. A potential discrepancy between the human and \textit{in silico} experiments:} As discussed earlier, the human experiments in \citet{russakovsky2015} used a few-shot learning design where the participants had access to 13 labeled examples from each class. However, it would be na\"ive to think that for many of the classes in ImageNet, especially for the more common classes, these labeled examples constituted the totality of the labeled examples the participants had encountered in their lifetime. This puts the self-supervised model at a disadvantage relative to the human participants, because for the model, the labeled examples seen during the few-shot learning evaluation are truly the only labeled examples it has encountered. It is difficult to precisely estimate the size of this effect, but our first scaling experiment suggests that this can potentially be a serious concern for that particular experiment (it is less of a concern for the OOD accuracy scaling experiment, as we have shown in Figure~\ref{robustness_fig} and in Table~\ref{geirhos_tab}), since we have observed that simply doubling the amount of labeled data available during few-shot evaluation (from $\sim$1\% to 2\% of the ImageNet training data) shaves off $\sim$2-3 orders of magnitude from our estimates (see Table~\ref{scaling_tab} and the green vs.~red fits in Figure~\ref{scaling_fig}). One way to alleviate this concern would be to use visual categories relatively unfamiliar to the human subjects, for example using only fine-grained categories in ImageNet (e.g.~different dog breeds etc.). \citet{russakovsky2015} indeed mention that human subjects had a higher error rate for fine-grained ImageNet categories and the self-supervised model would also be expected to have a higher error rate for these categories due to, e.g., the increased between-class similarity. In general, future work should try to use visual categories for which prior supervised exposure by human subjects can be more precisely controlled, thus allowing for a fairer comparison between humans and self-supervised models.
\vspace{-1em}
\section{Discussion}

To our knowledge, this work presents the first serious attempt at quantitatively comparing the sample efficiency of the current state-of-the-art self-supervised visual representation learning algorithms vis-\`a-vis humans in a complex, realistic visual object recognition task, namely ImageNet. Our main result is that current self-supervised learning algorithms would need on the order of millions to billions of years of human-like natural video data in order to achieve human-level accuracy and robustness on ImageNet. It is important to note that even under the most optimistic scenarios we have considered, our estimates remain quite large, often several orders of magnitude longer than a human lifetime. 

If this estimate is in the right ballpark (see the caveats above), one can broadly imagine two types of responses to it. Someone with empiricist sympathies might correctly point out that the full human sensorium is much richer than a video stream, that taking into account the embodied nature of human experience, in particular, might close the gap between humans and the self-supervised models \citep{hill2019,jacobs2019}, without having to significantly change the very generic nature of our models and learning algorithms. However, the burden of proof would be on the empiricist to quantitatively show that these additional inputs indeed contain enough extra information to close the sample efficiency gap we have found here. On the other hand, someone with more rationalist leanings might see the footprints of evolution operating in our estimates. The rationalist would consider it unlikely that additional inputs would significantly alter these estimates, hence the rationalist might instead advocate improving our current models and learning algorithms, possibly by incorporating more human-like inductive biases, to close the sample efficiency gap with humans. In this case, the burden would be on the rationalist to show that our current models and algorithms can indeed be significantly improved in this way. There is reason to be optimistic about this possibility given that recent years have seen remarkable progress in self-supervised learning algorithms (and also in model architectures). 

Visual object recognition represents only a single facet of human vision (important though it may be) and ImageNet classification represents only a single facet of visual object recognition. Future work should take into account a much broader range of tasks and perceptual capabilities in comparing the sample efficiency of humans vs.~self-supervised deep learning models.

In general, we hope that future work will run even larger scale, more realistic, and more ambitious versions of the scaling experiments we have performed here, addressing some of the caveats discussed in the previous section. Technical advances in machine learning increasingly allow us to pull off such ambitious experiments.

\section*{Acknowledgements}
My gratitude extends outward from me in expanding, reverberating waves: to those who made the datasets described in section~\ref{sec_datasets} available, to the developers of PyTorch, to the designers, manufacturers, and the maintainers of the GPUs on which these experiments were run (special gratitude to the HPC team at NYU), to the intellectual giants past and present on whose shoulders I stand, to the janitors who maintain the building where I work, to the workers in Vietnam and in Taiwan, who assembled the keyboard, the monitor, and the laptop I use to type this manuscript, to the workers in China who provide for their families by assembling chairs like the one I sit in while I type these words, to the baristas in NYC who give me my morning coffee, to the Colombian farmers who cultivate those coffee beans, to all the food workers who feed me and all the textile workers who clothe me.~To every single human being on this planet working tirelessly every day to make somebody else's life easier and happier.

\bibliography{natural_vision}
\bibliographystyle{apalike}

\section*{Appendix}
\subsection*{Training details}
The data augmentation pipeline used in the temporal classification algorithm was as follows:

\texttt{\footnotesize import torchvision.transforms as transforms \\
\\
train\_transforms = transforms.Compose([ \\
    transforms.RandomResizedCrop(224, scale=(0.08, 1.0)), \\
    transforms.RandomApply([transforms.ColorJitter(0.9, 0.9, 0.9, 0.5)], p=0.9), \\
    transforms.RandomGrayscale(p=0.2), \\
    transforms.RandomApply([GaussianBlur([0.1, 2.0])], p=0.5), \\
    transforms.RandomHorizontalFlip(), \\
    transforms.ToTensor(), \\
    transforms.Normalize(mean=[0.485, 0.456, 0.406], std=[0.229, 0.224, 0.225]) \\
    ])}

where \texttt{GaussianBlur} is a custom-written Gaussian blur transformation. Models were trained up to convergence. For the largest data size (with $\sim$23M frames), training took 12 epochs. For the optimization, we used the \texttt{Adam} optimizer with a learning rate of 0.0005 and a batch size of 256 in this case. All manual hyperparameter searches were performed only with models trained at the largest data size with respect to linear probe accuracy on the ImageNet training set.

The data augmentation pipeline used for the DINO algorithm consisted of two global crop augmentations and eight local crop augmentations. The global crop augmentations were as follows:

\texttt{\footnotesize import torchvision.transforms as transforms \\
\\
flip\_and\_color\_jitter = transforms.Compose([ \\
    transforms.RandomHorizontalFlip(p=0.5), \\
    transforms.RandomApply([transforms.ColorJitter(0.9, 0.9, 0.9, 0.5)], p=0.9), \\
    transforms.RandomGrayscale(p=0.2),\\
])\\
\\
normalize = transforms.Compose([\\
    transforms.ToTensor(),\\
    transforms.Normalize((0.485, 0.456, 0.406), (0.229, 0.224, 0.225)),\\
])\\
\\
First global crop: \\
global\_transform\_1 = transforms.Compose([ \\
    transforms.RandomResizedCrop(224, scale=(0.15, 1), interpolation=Image.BICUBIC),\\
    flip\_and\_color\_jitter, \\
    GaussianBlur(1.0),\\
    normalize,\\
])\\
\\
Second global crop:\\
global\_transform\_2 = transforms.Compose([\\
    transforms.RandomResizedCrop(224, scale=(0.15, 1), interpolation=Image.BICUBIC),\\
    flip\_and\_color\_jitter,\\
    GaussianBlur(0.1),\\
    Solarization(0.2),\\
    normalize,\\
])}

where \texttt{GaussianBlur} and \texttt{Solarization} are custom-written Gaussian blur and solarization transformations, respectively. The local crop augmentations were as follows:

\texttt{\footnotesize local\_transform = transforms.Compose([\\
    transforms.RandomResizedCrop(96, scale=(0.05, 0.15), interpolation=Image.BICUBIC),\\
    flip\_and\_color\_jitter,\\
    GaussianBlur(p=0.5),\\
    normalize,\\
])
}

Models were trained up to convergence using the \texttt{AdamW} algorithm with a learning rate of 0.0005, a weight decay coefficient of 0.0001, and a batch size of 556 (the largest batch size we could fit into 4 NVIDIA V100 GPUs). We employed half-precision training to speed up training and used gradient norm clipping with a threshold value of 0.3 to ensure stability during training.

\subsection*{Details related to the \citet{geirhos2021} OOD generalization benchmarks}
The 17 OOD generalization datasets from \citet{geirhos2021} involve the following image modifications/distortions/manipulations: \textit{edge, silhouette, cue-conflict, sketch, stylized, color, contrast, high-pass, low-pass, phase-scrambling, power-equalization, false-color, rotation, eidolonI, eidolonII, eidolonIII, uniform-noise}. Further details about each of these conditions can be found in \citet{geirhos2021}.

The 16 basic-level categories used in these OOD generalization experiments are: \textit{airplane, bear, bicycle, bird, boat, bottle, car, cat, chair, clock, dog, elephant, keyboard, knife, oven, truck}.
\newpage
\subsection*{Data points}

In Figure~\ref{scaling_fig} in the main text:

\texttt{\footnotesize --- Temporal Classification --- \\
finetuned with 1\%: \\
20.356, 20.356, 21.388, 31.688, 30.544, 29.692, 35.040, 35.486, 34.296, 42.958, 42.332, 42.426 \\
\\
finetuned with 2\%: \\
34.238, 29.392, 30.076, 45.036, 43.330, 42.334, 47.408, 48.498, 47.636, 54.304, 54.530, 55.044
\\
\\
linear probe (frozen): \\
25.426, 21.740, 24.654, 45.316, 45.032, 42.262, 52.620, 57.828, 53.482, 66.740\\
\\
--- DINO --- \\
finetuned with 1\%:\\
26.296, 30.252, 31.910, 33.178, 36.514, 35.748, 39.192, 38.032, 39.730, 43.658, 43.318, 44.068\\
\\
finetuned with 2\%:\\
37.350, 41.764, 43.544, 45.010, 48.512, 47.700, 50.732, 50.190, 50.340, 55.562, 55.548, 55.980\\
\\
linear probe (frozen): \\
44.798, 53.168, 51.676, 55.858, 60.188, 59.484, 62.044, 62.764, 60.802, 67.206\\
}

In Figure~\ref{robustness_fig} in the main text:

\texttt{\footnotesize --- Temporal Classification --- \\
finetuned with Geirhos practice images: \\
14.4640, 13.7262, 14.9921, 20.7911, 21.6308, 23.2634, 27.3493, 27.1969, 27.2851, 28.9340, 29.2668, 29.4755\\
\\
finetuned with Geirhos practice images + 2\% ImageNet: \\
17.0296, 15.2393, 17.7015, 24.4920, 25.0961, 26.3706, 28.2567, 28.2872, 28.0853, 32.0132, 30.9767, 31.3912 \\
\\
--- DINO --- \\
finetuned with Geirhos practice images: \\
24.7310, 21.7794, 20.5854, 23.4211, 23.3952, 24.1744, 24.3888, 23.2622, 25.1338, 25.7913, 25.5840, 26.1908 \\
\\
finetuned with Geirhos practice images + 2\% ImageNet: \\
25.2495, 24.0602, 22.7981, 25.0021, 25.1748, 24.0902, 25.0161, 26.2747, 26.8167, 28.1907, 27.5316, 27.0729\\
}

\begin{figure}
  \centering
    \includegraphics[width=1.0\textwidth, trim=0mm 0mm 0mm 0mm, clip]{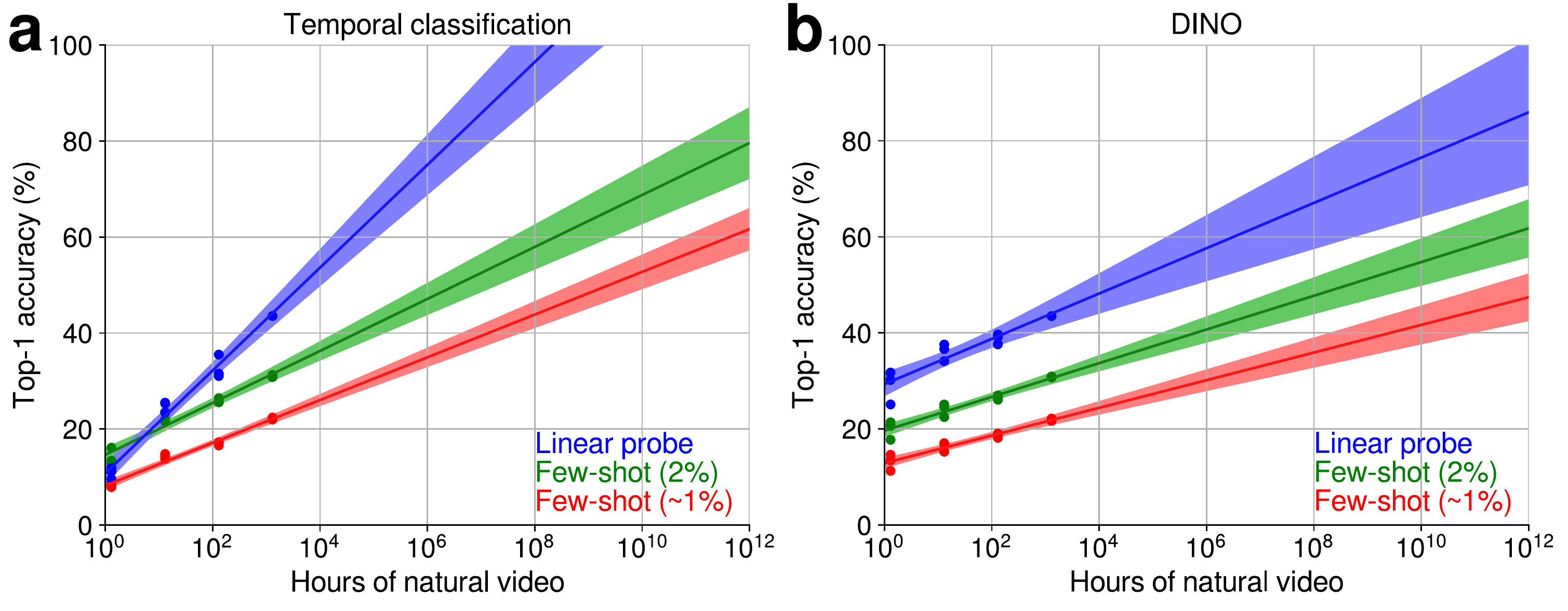}
  \caption{Top-1 validation accuracy on ImageNet as a function of the amount of natural video data used for self-supervised learning with the temporal classification (a) and DINO (b) algorithms.~Results with three different evaluation protocols are shown: linear probing (blue), few-shot learning with $\sim$1\% of the labeled training data (red), and few-shot learning with 2\% of the labeled training data (green).~Each dot represents a different run.~The straight lines are linear fits, the shaded regions represent 95\% confidence intervals around the linear fit.}
  \label{top1_fig}
\end{figure}

In addition, we show in Figure~\ref{top1_fig} below the scaling behavior of the top-1 accuracy on the clean ImageNet validation set both for the temporal classification and DINO algorithms.

\end{document}